# Design, Characterization, and Control of a Size Adaptable In-pipe Robot for Water Distribution Systems

Saber Kazeminasab
*Department of Electrical and Computer Engineering*
*Texas A&M University*
College Station, TX, USA
skazeminasab@tamu.edu

Ali Akbari
*Department of Biomedical Engineering*
*Texas A&M University*
College Station, TX, USA
aliakbari@tamu.edu

Roozbeh Jafari
*Departments of Biomedical, Computer Science and Electrical and Computer Engineering*
*Texas A&M University*
College Station, TX, USA
rjafari@tamu.edu

M. Katherine Banks
*College of Enginnering*
*Texas A&M University*
College Station, TX, USA
k-banks@tamu.edu

*Abstract*— Leak detection and water quality monitoring are requirements and challenging tasks in Water Distribution Systems (WDS). In-line robots are designed for this aim. In our previous work, we designed an in-pipe robot [1]. In this research, we present the design of the central processor, characterize and control the robot based on the condition of operation in highly pressurized environment of pipelines with the presence of high-speed flow. To this aim, an extreme operation condition is simulated with computational fluid dynamics (CFD) and the spring mechanism is characterized to ensure sufficient stabilizing force during operation based on the extreme operating condition. Also, an end-to-end method is suggested for power considerations for our robot that calculates minimum battery capacity and operation duration in the extreme operating condition. Finally, we design a novel LQR-PID based controller based on the system's auxiliary matrices that retains the robot's stability inside pipeline against disturbances and uncertainties during operation. The ADAMS-MATLAB co-simulation of the robot-controller shows rotational velocity with -4°/sec and +3°/sec margin around x, y, and z axes while the system tracks different desired velocities in pipelines (i.e. 0.12m/s, 0.17m/s, and 0.35m/s). Also, experimental results for four iterations in a 14-inch diameter PVC pipe show that the controller brings initial values of stabilizing states to zero and oscillate around it with a margin of $\pm 2°$ and the system tracks desired velocities of 0.1m/s, 0.2m/s, 0.3m/s, and 0.35m/s in which makes the robot dexterous in uncertain and highly disturbed environment of pipelines during operation.

## I. INTRODUCTION

Water Distribution Systems (WDSs) are critical infrastructures that transfer drinking water to consumers. The pipelines corrode over time and a small crack may let the contaminants enter the network which causes pollution that threatens people health. These contaminants can be introduced to water pipes by accidental or deliberate incidents. Therefore, timely measurement of water quality is required to ensure the safety of the potable water in WDS. Also, when leak occurs in the network, it should be localized and fixed as soon as possible as water loss accounts for large portion of purified water in water utilities. Water loss for the US is reported at 15%-25% [2] and 20% for Canada [3]. The pipelines are usually lengthy and traditional methods for quality monitoring and leak detection are not efficient, and they depend on technicians' experience.

The solution to this issue is mobile sensors. They can go into inaccessible parts of the network and measure the concentration of target analytes in water. Water flow carries these small sensor modules inside the pipeline [4]. However, WDSs have complicated configurations and it is highly probable that an operator loses the manually-controlled sensor module in the network during an operation. Thus, it is necessary to design sensor modules in a way that they have controlled motion in the network. The controlled motion enables the robot to move inside pipelines and tackles disturbances and uncertainties as the conditions in pipelines are highly uncertain and the presence of flow results in disturbances for the robot motion.

In-pipe robots can carry these sensors inside the pipeline with controlled and reliable motion. They have attracted researcher's attention during the past few years due to the miniaturization of mechatronic components. Several robotic systems have been proposed to perform specific tasks (e.g., leak detection, quality monitoring, visual inspection, etc.) in the pipeline; however, they suffer from several limitations. These robots are actuated either by pneumatic actuators [5] or electrical actuators [6]. The pneumatic actuated robots need special mechanisms to transform the actuator power to the mechanical components which makes their motion slow. Another issue is related to varying size of the pipelines that require the robots to adapt to different pipe diameters. Some robot mechanisms just adapt to a specific size [7] while others can change their outer diameter based on pipe diameter [8].

Some robots are powered with cable [9] and they have this length limitation thus battery powered robots are preferred. Environmental pressure is another issue for in-pipe robots. The operation environment of the in-pipe robots are pressurized pipelines in which a water flow is present which puts disturbances on the robot. Finally, the uncertain and complicated configuration of the pipelines requires the robots to intelligently navigate through the pipes. Technical problems in the field of in-pipe robots can be summarized as:

- The robots' mechanisms are complicated and interfere with water flow.
- Current mechanisms for in-pipe robots are not maneuverable in special configurations of pipelines like bends and junctions.
- Power considerations is not well addressed in current robots.
- The robots designed for large pipe sizes are limited in terms of motion efficiency to negotiate complicated configurations of pipelines, and also inspection distance [10]. They need special equipment and launching system for inserting them into the pipeline and retrieving them from the pipeline after operation.

Considering the mentioned challenges, we provided a methodology for designing the in-pipe robots in [11]. The contributions of this paper are as follows:

- We further develop our previously designed robotics' body [1] here and present the central processor that enables the robot for both quality monitoring and leak detection.
- The spring-mechanism is characterized based on real application condition. An extreme operation condition is simulated with computation fluid dynamic (CFD) work to have a **fully automotive system** in pipelines.
- Toward fully automotive system, we also provide an end-to-end method for power considerations that calculate the minimum battery capacity and operation duration considering the extreme operation condition.
- We design a controller algorithm that controls velocity of the robot and stabilizes it against uncertainties by decoupling the dynamic equations and representing the stabilizing states and deriving the system's auxiliary matrices.
- The control algorithm along with the robot is simulated with ADAMS MATLAB co-simulation.
- We also conduct an experiment and evaluate the controller performance in a pipe.

The remainder of this paper is organized as follows. In section II, the robot design and its components are presented and then modeled. In section III, the robot is characterized based on an extreme condition that is modeled with a CFD simulation. In section IV, a control algorithm is proposed and its performance is analyzed with simulation and experiments in section V. The paper is concluded and future works are presented in section VI.

## II. ROBOTIC DESIGN AND MODELING

### A. Robot Design

The robot is composed of:

- One central processor.
- Three arm modules.
- Three actuator modules.

In Fig. 1, the overall view of the robot and its components are shown. The central processor hosts a sensing unit, a data processing unit, a control unit, and a power unit. Two components compose the central processor. One of them is designed to locate the sensor modules for measurements of water parameters. We call it the sensing part for brevity. The sensing part is designed to host miniaturized sensor modules that need water samples for measurements (e.g. water quality monitoring sensors, see Fig. 2b). A micro-pump system provides and circulate water into sensor modules through water inlets and outlets (see Fig. 2c). Acoustic sensors can be used in the sensing part for leak detection in WDS. The electronic embedded system designed for the robot, controls the motion of the robot, processes sensor measurements, communicate wirelessly with base station. The embedded system (PCB) is located inside the other part of the central processor that we call it control part. The power supply which is a compact battery is located below the control part in a case (see Figs. 2a and 2b). This battery provides power for all parts of the system during operation.

Three arm modules are anchored on the central processor with 120° angles.

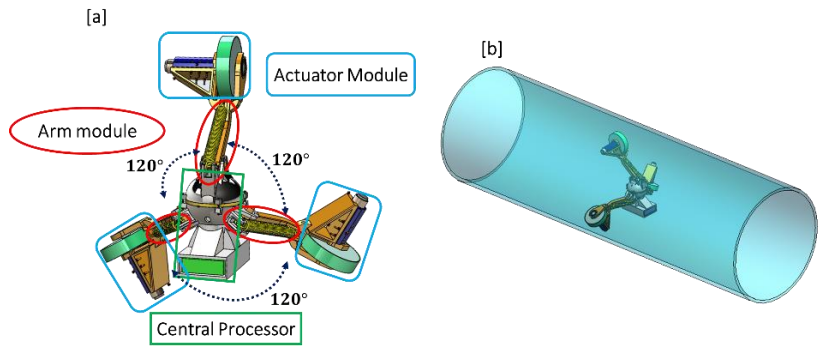

Fig. 1: CAD design for the proposed in-pipe robot. [a] Front view. [b] Robot in pipe.

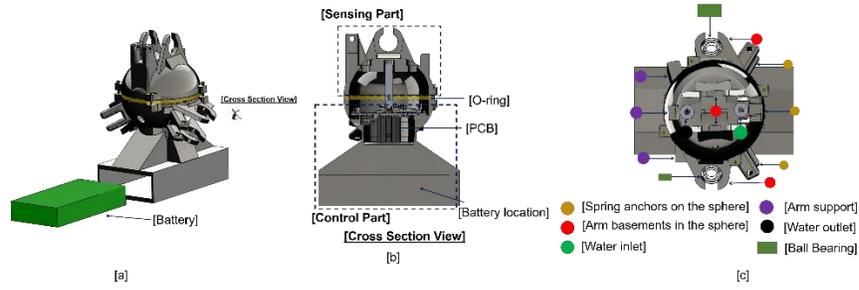

Fig. 2. The central processor [a] The central processor CAD design. [b] Cross section view of the central processor. [c] The above view of the central processor and water samples provider mechanism.

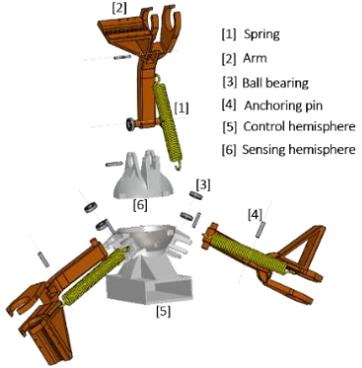

Fig. 3. Exploded view of the arm modules in the robot.

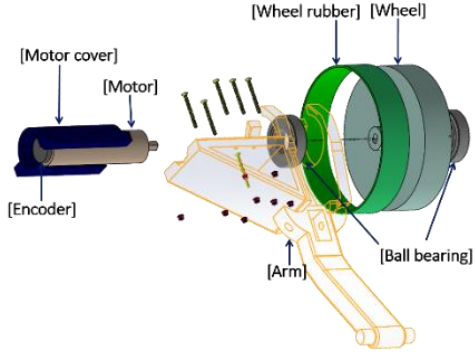

Fig. 4. Exploded view of the actuator module.

Fig. 3, shows the exploded view of the arm modules, the special shape of the arms prevents collision of the spring with it during extension and retraction. The spring causes the wheels at the actuator module (see Fig. 4) to press the pipe wall during operation (see Fig. 1b). The press force enables the robot, stabilize itself, and moves inside the pipe.

The details of the design and fabrication can be found in our previous work [1]. Also, the details of the robot components, the seal mechanism, the health issues consideration, and the controllability and observability of the robot are discussed in detail in [1]. In Fig. 5, the robot prototype is shown. The components are fabricated with acrylonitrile butadiene styrene (ABS) with 3D printing.

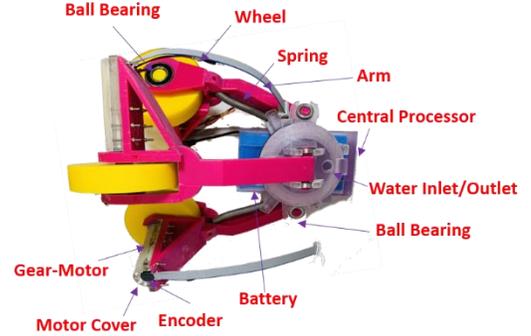

Fig. 5. Photograph of the robot prototype.

*B. Robot Modeling*

The robot in a pipe is shown in Fig. 6. $D$ is pipe diameter, $\tau_1$, $\tau_2$, and $\tau_3$ are motor torques. $mg$ is the robot weight. $F_d$ is drag force acting on robot due to robot and flow velocity difference [12]. The robot's equations of motion can be written with Lagrangian energy function that is:

$$\mathcal{L} = T - V \quad (1)$$

Where $T$ is the kinetic energy of components and $V$ is the potential energy of the robot's components. The governing equations can be written as follows:

$$\frac{d}{dt}\left(\frac{\partial \mathcal{L}}{\partial \dot{x}}\right) - \frac{\partial \mathcal{L}}{\partial x} = J^T . \mathcal{F} \quad (2)$$

where, $t$ is time, $x$ is the system states, $J$ is the Jacobian transformation matrix, and $\mathcal{F}$ is the external forces acting on the robot. Derivation of the system equations is explained in our previous work in detail [12]. We present the systems governing equations in a general form of:

$$\dot{x} = f(x, u) \quad (3)$$

In (3), $u = \begin{bmatrix} \tau_1 \\ \tau_2 \\ \tau_3 \end{bmatrix}$ is the input vector to the robot.

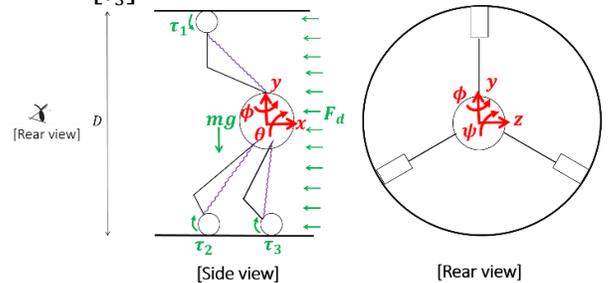

Fig. 6. Free body diagram of the robot in pipe.

- Motor Modeling

The actuators here are gear-motors. Hence we analyze the electromechanical behavior of them in this part. We have:

$$\frac{v_{co}}{L_m} - \frac{v_e}{L_m} - \frac{R_m}{L_m} i_m = \frac{di_m}{dt} \quad (4)$$

Where $R_m$ is the terminal resistance, $L_m$ is the terminal inductance. $i_m$ is the current through motor. $v_m = K_v \dot{\vartheta}$ is the back EMF of the motor ($\dot{\vartheta}$ is motor shaft angular velocity). $T_m = K_v i_m$ is the mechanical torque at the motor shaft. $v_{co}$ is the input voltage on the motor and our control variable. We have:

$$\frac{n^2}{I_l + n^2 I_R} T_m = \ddot{\vartheta} \quad (5)$$

$n$ is the reduction ration of the gear, $I_l$ and $I_R$ are load inertia and rotor inertial, respectively.

## III. ROBOT CHARACTERIZATION

The robot operates in pipeline which is a pressurized environment with a high-speed water flow. The robot presses the pipe wall to stabilize itself. The operation condition affects the robot parameter and these parameters need to be modeled based on real operation conditions. To this aim, we simulated the robot in pipe and the flow around it with computational fluid dynamics (CFD) work in SolidWorks. In the CFD work, an extreme condition is considered where the robot is in the extended situation (outer diameter is 22 inch) and the flow and the robot directions are opposite (see Fig. 7). The pipe wall is defined in Fig. 7. The robot velocity is 50cm/s and the flow velocity is 70cm/s. The colored lines show the velocity in the pipe and around the robot. In this scenario, the maximum drag force was computed to be around 18N. In Fig. 7, the simulation environment and the velocity counter around the robot in the simulation is shown.

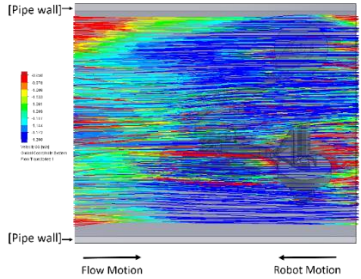

Fig. 7. CFD work In SolidWorks for maximum drag force computation.

We characterize the spring mechanism and also analyze power supply based on the results of the CFD work.

### A. Spring Stiffness Characterization

The operating environment is highly pressurized. If the robot stocks during operation, the network should be shut down. then an access point needs to be created on the location of the stocked robot and the robot is retrieved. Retrieving the stocked robot is a costly and cumbersome task. Hence, the robot needs to be fully automotive in pipelines during operation.

We formulate the problem of crawling our robot inside the pipe based on the spring mechanism. This work addresses the issue of controlled motion in the pressurized environment. The objective of the spring characterization is to find a proper value for stiffness that ensures reliable motion of the robot. In this section, the problem is formulated based on the robot geometry and maximum traction force that **each wheel** needs to provide (i.e., 6N).

A wheeled- robot has controlled motion if the wheels have absolute rolling. Hence, the maximum traction force generated at the contact area needs to be less than the maximum friction force between the pipe and the wheel:

$$F_{T(max)} \leq f_{s(max)} \quad (6)$$

where $F_{T(max)}$ is the maximum traction force and $f_{s(max)}$ is the maximum friction force. The robot is considered symmetric and the wheels below its geometric center bear the robot weight and the weight of the robot increases the contact force, $F_N$ (see Fig. 8a). Hence, the required spring for them is smaller. To calculate the maximum spring force, we consider the wheel that is above the center of mass. Also, we consider all the robot's weight is considered at one wheel static force analysis for it (Fig. 8a). Also, all the weight of the robot is assumed at the pipe wheel contact. In Fig. 8b, the free body diagram of the arm is shown. If the central processor remains at the center of the pipe (which is needed), the static force analysis on point $O$ can be written as:

$$\sum M_0 = 0 \rightarrow (mg - F_N) a \cos\beta = f_s \cdot H + F_{Spring} \chi_{Spring} \quad (7)$$

Based on $OAB$ triangle, we can write:

$$\beta = \alpha + \left(\frac{\pi}{2} - \theta\right) \quad (8)$$

$$\alpha = \sin^{-1}\left(\frac{t}{a}\cos\theta\right) \quad (9)$$

Plugging (9) into (8):

$$\beta = -\theta + \sin^{-1}\left(\frac{t}{a}\cos\theta\right) + \frac{\pi}{2} \quad (10)$$

On the other side, we have:

$$\beta = \sin^{-1}\left(\frac{H}{L}\right) \quad (11)$$

Considering (10) and (11), we have a nonlinear trigonometric equation that $\theta$ (see Fig. 8b) is calculated based on the pipe radius, $H$. We also can write:

$$\chi_{Spring} = t \cos\theta \quad (12)$$

Plugging (8), (9), (10), (12) into (7), we can write the spring force, $F_{Spring}$, as:

$$F_{Spring} = \frac{1}{t\cos\theta}\left((F_N - mg)a\cos\left(\theta + \sin^{-1}\left(\frac{t}{a}\cos\theta\right)\right) - f_s H\right) \quad (13)$$

The springs are linear in which the force is proportional to its displacement. Hence, based on the geometry of the anchor points (**A** and **B** in Fig. 8b), the initial length of the spring has no pretension at $\theta = 0$, and is calculated as $\sqrt{(t+\cos\beta)^2 + (a\sin\beta)^2} \cos\theta$. The length of the spring at $\theta > 0$ is calculated as $\sqrt{(t+\cos\beta)^2 + (a\sin\beta)^2}$. If $K$ is the spring stiffness, $F_{Spring}$ is calculated as:

$$F_{Spring} = K(\sqrt{(t+\cos\beta)^2 + (a\sin\beta)^2})(1-\cos\theta) = KU(\theta) \quad (14)$$

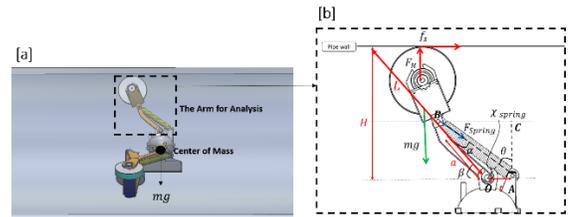

Fig. 8: The problem geometry for spring stiffness characterization.

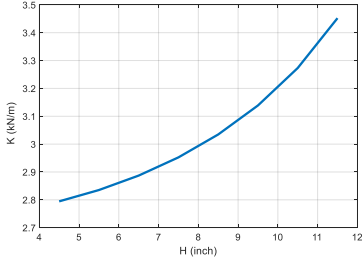

Fig. 9. Stiffness value calculation for pure rolling of wheels during operation.

The value for $a$=103 mm and $t = 36$mm. Plugging (14) in to (13), $K$ is calculated as:

$$K = \frac{1}{t\cos\theta(U(\theta))}\left((F_N - mg)a\cos\left(\theta + \sin^{-1}(\frac{t}{a}\cos\theta)\right) - f_s H\right) = G(\theta) \quad (15)$$

The appropriate value for $K$ is the maximum of the (15), Hence:

$$K = max(G(\theta)) \quad (16)$$

The maximum traction force computed with the CFD work in SolidWorks is considered for $f_s$ that is 6N based on (6). Also, for $H$, the values that the robot can cover are considered (4.5inch≤ $H$ ≤11inch). The friction between pipe and wheel contact is considered dry coulomb. Hence, the relation between $F_N$ and $f_{s(max)}$ is:

$$f_{s(\max)} = \mu_s F_N \quad (17)$$

In (17), $\mu_s$ is the friction coefficient and is considered 0.8 based on [13]. Hence, the value for $F_N$ is considered 7.5N. In Fig. 9, different values for $K$ that ensures pure rolling for wheels in all positions of arm and different pipe diameters are shown. The maximum value in Fig. 9 is selected, hence in all pipe sizes, the pressing force is sufficient for pure rolling (i.e. controlled motion) of the wheels.

The value range computed for $K$ is large, hence, the anchoring points are designed to bear the spring forces. As can be seen in Fig. 3, the springs are anchored on the central processor and arms with metal pins. The pins are thin 416 alloy stainless steel with a 0.2-inch diameter and 0.9-inch length. Also, the anchor bases are designed to be strong enough. So far, a simple modular mechanism is designed for the robot to press the pipe wall and ensure sufficient contact force for the wheels. We assumed that the central processor remains at the center of the pipe during operation. To make it possible, we design a control algorithm to keep it at the center of the pipe, later in this paper.

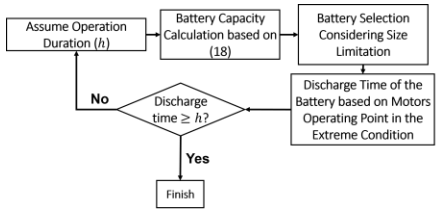

Fig. 10: Recursive approach to calculate minimum battery capacity and operation duration based on the operation conditions of the robot.

*B. Power Considerations*

The power considerations for battery-powered robots is not well addressed in the literature. In this section, we provide a novel recursive method to calculate the **minimum battery capacity and operation duration** for the robot based on conditions in real applications. The main power consumers in our robot are gear-motors. (Other components like wireless communication system consume negligible power compared to gear motor). Hence, gear-motors are considered for power profiling. In our proposed method, an operation duration, $h$, is assumed. Based on the selected gear-motors' properties (power and nominal voltage), and also the number of motors in the robot, the minimum battery capacity is calculated as follows:

$$C = \frac{3P.h}{V_n} \quad (18)$$

In (18), $C$(A.h) is the battery capacity, $h$(hour) is operation duration, $P$ and $V_n$ are the motors' power and nominal voltage, respectively. Factor three in (18) accounts for number of the motors in our robot. Afterwards, a battery, considering **size limitations in the robot** is selected. The discharge time of the battery is defined based on the battery characteristics (defined in its datasheet) and the drawn electrical current by the gear-motors. The amount of the drawn current in this method depends on the maximum torque the gear-motors are required to provide in the extreme condition (i.e. calculated in the CFD simulation). The process is repeated until the operation duration and the discharge time are approximately equal (see Fig. 10). For our robot, the minimum battery capacity is calculated 15A.h and the operation duration of the robot is 3 hours.

IV. CONTROLLER DESIGN

The under-actuated design of the robot is inherently unstable. There are uncertainties in pressurized environment of pipelines. Therefore, it is highly important for the robot to stabilize itself and retain it during operation. To this aim, we design a novel controller that stabilizes the robot during operation and tackles uncertainties and disturbances of operation environment and also tracks desired velocity.

*A. LQR Stabilizer Controller*

In the stabilization problem, the objective is to keep the central processor at the center of the pipe and keep it there against the weight of the robot and disturbances [12]. To this aim, we decoupled the dynamic equations of the system, (1) (i.e. $\dot{x} = f(x,u)$), into two sets: One set relates the input vector, $u$, to the robot odometry (i.e. $\dot{x}_1 = f_1(x_1, u)$, where $x_1$ is robot's odometry state vector). The other set relates the input vector to orientation of the robot (i.e. $\dot{x}_2 = f_2(x_2, u)$ where $x_2 = $

$[\phi \ \dot{\phi} \ \psi \ \dot{\psi}]^T$). $x_2$ are so-called stabilizing states that need to converge to zero and be kept at that point during operation. In section III.A, we characterized the spring-mechanism as the wheels have pure rolling during motion, hence, $\theta$ (see Fig. 6) would not change and we don't consider $\theta$ as a stabilizing state. We linearized $f_2(x_2, u)$ equilibrium point that is $\phi = \dot{\phi} = \psi = \dot{\psi} = 0$ and derived system's auxiliary matrices (i.e. $A_2 = \frac{\partial f_2}{\partial x_2}|_{x_2=0}$ and $B_2 = \frac{\partial f_2}{\partial u}|_{u=0}$ ). $C_2 = \begin{bmatrix} 0 & 1 & 0 & 0 \\ 0 & 0 & 1 & 0 \end{bmatrix}$ is the system's auxiliary output matrix. The system's auxiliary representation in state-space form is:

$$\dot{x}_2 = A_2 x_2 + B_2 u \quad (19)$$
$$y_2 = C_2 x_2 D_2 u \quad (20)$$

where $D_2$, feedforward matrix is zero in our system. Following onwards, we will design a state feedback controller based Linear Quadratic Regulator (LQR) approach [14]. To this aim, we define a cost function

$$J(K) = \frac{1}{2}\int_0^\infty [x_2^T(t)Qx_2(t) + u(t)^T Ru(t)]dt \quad (21)$$

Where $Q$ weights the stabilizing states and $R$ weights the input vector. The cost function depends on $K$ and $J(K)$ is minimized with the value of $K$ that is computed as:

$$K = R^{-1} B_2^T P \quad (22)$$

Matrix $P$ is computed with the algebraic Riccati equation:
$$-PA_2 - A_2^T P - Q + PB_2 R^{-1} B_2^T P = 0 \quad (23)$$

The gain matrix is then used to calculate the LQR stabilizer's output, $u_2$, that is:

$$u_2 = -Kx_2 \quad (24)$$

(24) presents the architecture of the state feedback controller that keeps $x_2$ at equilibrium point during operation.

*B. Velocity Controller*

A PID-based controller is designed that enables the robot to track the desired linear velocity along the pipe axis. In a straight path where the robot has linear motion, the velocity of wheels are approximately equal and the robot velocity, $v$, is computed with this:

$$v \approx \frac{2\pi R}{N[T_c]_i} \quad (25)$$

where $N$ is the number of pulses per wheel turn that incremental encoders generate. $R$ is the wheel radius. $T_c$ is the time between two consecutive pulses and $i = 1,2,3$. In our velocity control problem, desired linear velocity for the robot, $V_d$, is converted to the desired wheels' reference velocity (see Fig. 11) and the PID controllers track the reference velocities on each wheel [12].

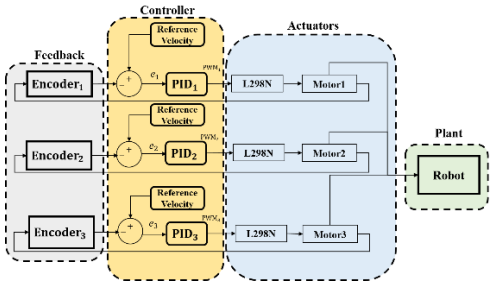

Fig. 11. The Velocity controller algorithm for the robot.

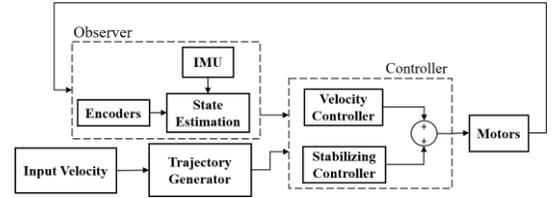

Fig. 12. The controller to stabilize the robot and track a desired velocity.

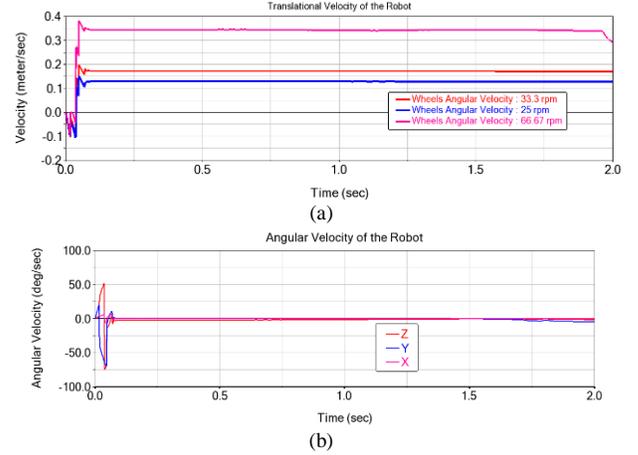

Fig. 13. Controller functionality in the straight pipe in ADAMS-MATLAB co-simulation. a) The robot linear velocity. The legend shows the wheels' angular velocity in revolutions per second (RPM) in each desired velocity. b) The robot's rotation velocity around x, y, and z axes in the situation the desired linear velocity is 0.35m/s.

*C. Combined LQR-Velocity Controller*

So far, we developed two controllers; the LQR controller stabilizes the robot during operation and the velocity controller enables the robot to track the desired velocity. Since the robot needs to have both stabilization and controlled motion, the developed LQR stabilizer and velocity controller are combined and shown in Fig. 12. The observer in Fig. 12, provides information about the orientation and odometry of the robot. Input velocity is the desired linear velocity along the pipe axis for the robot, $V_d$. The Trajectory generator computes the desired values for the robot's stabilizing states and wheels' angular velocities.

V. SIMULATION, EXPERIMENTAL ANALYSIS AND DISCUSSION

*A. Simulation Analysis*

To evaluate the controller performance, we modeled the robot in ADAMS, dynamic simulation software, and co-simulated it with the designed controller in MATLAB 2020. In this way, the controller is implemented in MATLAB Simulink toolbox and connected to the ADMAS in real-time during the simulation. In Fig. 13, the simulation results are shown. The coordinate system in the simulations is the same as shown in Fig. 6. It should be mentioned that in this simulation, the wheels are not in contact with the pipe wall at the beginning of the simulation. Hence, there is a fluctuation in the linear velocity of the robot (Fig. 13a) and angular velocities of the robot (Fig. 13b) for the first 0.1 seconds of simulation. After the wheels contacted the pipe wall, the fluctuations converged to zero. In Fig. 13a, the legends define the wheel's angular velocities in revolutions per second (RPM). The robot moves in the pipe axis with desired velocities; 0.12m/s, 0.17m/s, and 0.35m/s with stabilized reliable motion (i.e. angular velocities are zero). In Fig. 13b, the angular velocities of the robot along each axis (x, y, and z) are shown for the case the desired robot velocity is 0.35m/s. In Fig. 13b, the angular velocity of the robot on each axis ranges between -4 °/second to +3 °/sec. We also considered external disturbances in our simulations to evaluate the robustness of the controller in the presence of flow. The simulation results show the feasibility of the controller for the straight path as the robot can track a specific velocity in the pipe with balanced orientation.

*1) Experiment Setup*

To test the performance of the developed control strategy with experiments, we developed a test-bed that is shown in Fig. 14. The 6-DOF inertial measurement unit (IMU) BMI160 from Bosch Sensortec that includes accelerometer and gyroscope is located at the center of central processor and connected to the microcontroller unit (MCU) with inter-integrated circuit (I2C) protocol. The MCU is Arduino Mega2560 based on ATmega2560 microcontroller with 16MHz clock speed. Three incremental encoders with six channels are connected to the MCU through general purpose input output (GPIO) pins. Two dual full bridge motor driver boards, L298N from STMicroelectronics are connected to the MCU. Each driver controls two motors. The motor drivers control direction and velocity of the gear motors based on GPIO and Pulse Width Modulation (PWM) signals from the MCU. The experiment's parameters are defined in Table I. The gear-motors are powered with a 12V and 15A.h super lithium-ion battery. The MCU and IMU are powered with PC through USB cable. The motor drivers are powered with both the MCU and the battery. For our experiments, we locate the system in a 14-inch diameter SCH.40 PVC 1120 pipe. The schematic of motion is depicted in Fig. 14b.

Since the IMU outputs are noisy and have a DC baseline, $\phi$ and $\psi$ are computed with Mahony complementary filter [15]. The robot's linear velocity is computed with (25). The robot velocity along with Mahony filter data are monitored in real-time with MATLAB support package for Arduino hardware in MATLAB 2020.

*2) Experimental Results and Discussion*

The control algorithm on the robot is tested with four iterations. The robot is located in the pipe manually and when the arms are in contact with the pipe wall, the tests start. In Fig. 15, the results for the stabilizing states and the velocity of the robot are presented. In Fig. 15a, the robot velocity and in Figs. 15b and 15c, the robot-controller performance for $\phi$ and $\psi$ for each iteration are shown (e.g. the blue curve in Figs. 15a, 15b, and 15c represents the robot's velocity and stabilizing parameters for iteration 1.). At the beginning of motion, the robot velocity is zero and reaches the desired velocity and tracks the velocity for the rest of the motion. The values for $\phi$ and $\psi$ are non-zero in the iterations due to the initial orientation of the central processor. In iteration 1, $\phi$ is -4° and $\psi$ is -3° at the time the experiment starts. These values converge to zero and fluctuate around it with a margin of $\pm 2°$ in less than two seconds. The robot velocity reaches to 0.1m/s in around two seconds since the motion starts. In iteration 2, the initial value for $\phi$ and $\psi$ are -14° and -11°, respectively, and the desired linear velocity is 0.2m/s. The controller zeros the stabilizing states in two seconds and reaches to 0.2m/s velocity in around three seconds. For iteration 3, the initial value for $\phi$ and $\psi$ are -9° and +5°, respectively, and the desired linear velocity is 0.3m/s. $\phi$ and $\psi$ converge to zero in one second and the velocity reaches the desired value (i.e. 0.3m/s) within in two seconds and a half. The stabilizing duration is again around one second in iteration 4, however, it took around five seconds since the robot reaches the desired velocity of 0.35m/s.

The experiments prove that the under-actuated design of the robot with a robust controller is capable of providing reliable motion in pipes. In this experiment, the controller stabilizes the robot against its weight and locates the central processor in the center of the pipe. In real applications, there are a lot of uncertainties that the robot needs to negotiate. For example, due to sediments in pipes, the internal shape of the pipe is not circular. With the stabilizing $\phi$ and $\psi$, the central processor remains at the center of the pipe. Being at the center, the cross-section area of the robot facing flow becomes symmetric in which reduces disturbing torques on the robot during operation and decreases power consumption.

Table I. CONTROL PERFORMANCE EVALUATION EXPERIMENT COMPONENTS.

| Component | Description |
| --- | --- |
| IMU | Bosch Sensortec BMI160 breakout |
| Encoder | 6 channels ENX EASY 16 |
| Motor | Maxon DCX22S, 12V |
| Gear Type | Ultra-performance GPX22UP, 26:1 |
| MCU | Arduino Mega2560 |
| Motor Driver | L298N |
| Power Supply | Super lithium-ion battery |

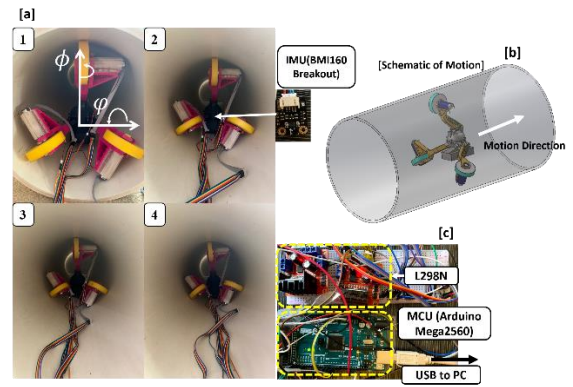

Fig. 14. Experiment setup for the controller. [a] The robot motion in pipe. [b] Schematic of motion. [c] Control and drive components in the experiment.

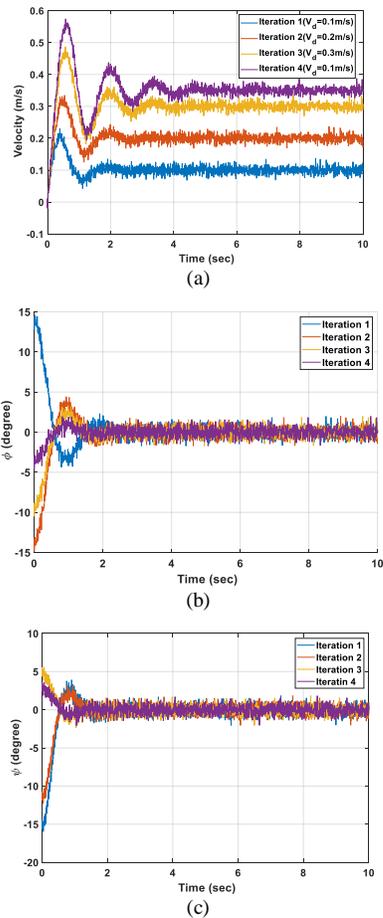

Fig. 15. Velocity and stabilizing states in the experiment. (a) Linear velocity. b) $\phi$ (degree). c) $\psi$ (degree).

## VI. CONCLUSION AND FUTURE WORKS

In this research, an under-actuated battery-powered in-pipe robot design is presented that operates in Water Distribution Systems (WDS) while the system remains in service. The adjustable arms equipped with a spring mechanism enable the robot to adapt to different pipe sizes. The modular and novel design of the sensing part in the central processor which is equipped with a micro-pump system makes it suitable for leak detection and water quality monitoring. Since the robot operate while the network is in service, the robot must be fully automotive in a highly pressurized environment in presence of high-speed flow. To this aim, we simulated an extreme operation condition in the pipeline with computation fluid dynamic (CFD) work in which the robot moves with a velocity of 0.5m/s in the opposite direction of a flow with 0.7m/s velocity in a 413kPa pressure environment. The spring mechanism is characterized to ensure sufficient crawling force for a fully automotive system based on the extreme operating condition. Also, an end-to-end method is presented to calculate the minimum battery capacity needed for the desired operation duration based on the operation condition and size limitations in the system. For our robot, a 15A.h battery capacity is needed for 3 hours operation duration. After deriving the system's auxiliary matrices for stabilization, an LQR-assisted velocity control algorithm is proposed to retain the robot's stability during the operation against disturbances and uncertainties in the pipeline and tracks the desired velocity. The proposed controller is evaluated with simulation in ADAMS MATLAB co-simulation. Also, an experiment is conducted to this aim. Four iterations of the experiment show the controller can stabilize the robot and track the desired velocities 0.1m/s, 0.2m/s, 0.3m/s, and 0.35m/s. The under-actuated design along with the proposed controller makes the robot a perfect option to operate in highly uncertain environments of the pipeline with huge disturbances at high speeds.

We plan to design multiple control phases for the robot. Also, a wireless communication system for the robot.

REFERENCES

[1] S. Kazeminasab, M. Aghashahi, and M. K. Banks, "Development of an Inline Robot for Water Quality Monitoring," in 2020 5th International Conference on Robotics and Automation Engineering (ICRAE), 2020, pp. 106–113, doi: 10.1109/ICRAE50850.2020.9310805.

[2] A. L. Vickers, "The future of water conservation: Challenges ahead," J. Contemp. Water Res. Educ., vol. 114, no. 1, p. 8, 1999.

[3] E. Canada, "Threats to water availability in Canada." National Water Research Institute Burlington, 2004.

[4] R. Wu et al., "Self-powered mobile sensor for in-pipe potable water quality monitoring," in Proceedings of the 17th International Conference on Miniaturized Systems for Chemistry and Life Sciences, 2013, pp. 14–16.

[5] K. Miyasaka, G. Kawano, and H. Tsukagoshi, "Long-mover: Flexible Tube In-pipe Inspection Robot for Long Distance and Complex Piping," in 2018 IEEE/ASME International Conference on Advanced Intelligent Mechatronics (AIM), 2018, pp. 1075–1080.

[6] Y. Qu, P. Durdevic, and Z. Yang, "Smart-Spider: Autonomous Self-driven In-line Robot for Versatile Pipeline Inspection," Ifac-papersonline, vol. 51, no. 8, pp. 251–256, 2018.

[7] M. Kamata, S. Yamazaki, Y. Tanise, Y. Yamada, and T. Nakamura, "Morphological change in peristaltic crawling motion of a narrow pipe inspection robot inspired by earthworm�s locomotion," Adv. Robot., vol. 32, no. 7, pp. 386–397, 2018.

[8] R. Tao, Y. Chen, and L. Qingyou, "A helical drive in-pipe robot based on compound planetary gearing," Adv. Robot., vol. 28, no. 17, pp. 1165–1175, 2014.

[9] A. A. Bandala et al., "Control and Mechanical Design of a Multi-diameter Tri-Legged In-Pipe Traversing Robot," in 2019 IEEE/SICE International Symposium on System Integration (SII), 2019, pp. 740–745.

[10] "PureRobotics® - Pipeline Inspection System - Pure Technologies - Pure Technologies." https://puretechltd.com/technology/purerobotics-pipeline-inspection-system/ (accessed Nov. 03, 2020).

[11] S. Kazeminasab, M. Aghashahi, R. Wu and M. K. Banks, "Localization Techniques for In-pipe Robots in Water Distribution Systems," 2020 8th International Conference on Control, Mechatronics and Automation (ICCMA), Moscow, 2020, pp. 6-11, doi: 10.1109/ICCMA51325.2020.9301560.

[12] S. Kazeminasab, R. Jafari, and M. K. Banks, "An LQR-assisted Control Algorithm for an Under-actuated In-pipe Robot in Water Distribution Systems," 2021, In press.

[13] "Material Contact Properties Table.", [Online]. Available: http://atc.sjf.stuba.sk/files/mechanika_vms_ADAMS/Contact_Table.pdf. [Accessed: 14-Nov-2020]

[14] E. V. Kumar and J. Jerome, "Robust LQR controller design for stabilizing and trajectory tracking of inverted pendulum," Procedia Eng., vol. 64, pp. 169–178, 2013.

[15] R. Mahony, T. Hamel, and J.-M. Pflimlin, "Nonlinear complementary filters on the special orthogonal group," IEEE Trans. Automat. Contr., vol. 53, no. 5, pp. 1203–1218, 2008.